# Multi-class real-time crash risk forecasting using convolutional neural network: Istanbul case study

Behnaz Alafi[a], Saeid Moradi[b]

[a]*Faculty of Electrical Engineering, Istanbul Technical University, Istanbul, Turkey*
[b]*Faculty of Civil Engineering, Istanbul Technical University, Istanbul, Turkey*

**Abstract**

The performance of an artificial neural network (ANN) in forecasting crash risk is shown in this paper. To begin, some traffic and weather data are acquired as raw data. This data is then analyzed, and relevant characteristics are chosen to utilize as input data based on additional tree and Pearson correlation. Furthermore, crash and non-crash time data are separated; then, feature values for crash and non-crash events are written in three four-minute intervals prior to the crash and non-crash events using the average of all available values for that period. The number of non-crash samples was lowered after calculating crash likelihood for each period based on accident labeling. The proposed CNN model is capable of learning from recorded, processed, and categorized input characteristics such as traffic characteristics and meteorological conditions. The goal of this work is to forecast the chance of a real-time crash based on three periods before events. The area under the curve (AUC) for the receiver operating characteristic curve (ROC curve), as well as sensitivity as the true positive rate and specificity as the false positive rate, are shown and compared with three typical machine learning and neural network models. Finally, when it comes to the error value, AUC, sensitivity, and specificity parameters as performance variables, the executed model outperforms other models. The findings of this research suggest applying the CNN model as a multi-class prediction model for real-time crash risk prediction. Our emphasis is on multi-class prediction, while prior research used this for binary (two-class) categorization like crash and non-crash.

*Keywords:* Real-time crash risk prediction; Traffic safety; Artificial neural network; Deep Learning; ROC curve, Multi-class Prediction

## 1. Introduction

Traffic flow has expanded rapidly and collisions have become more prevalent, traffic safety has become a critical concern for rural roads and urban expressways. Despite several initiatives, expressway safety remains a key cause of concern. Every year, around 1.25 million people are killed in road accidents, according to the World Health Organization's Global Status Report on Road Safety (Ren et al., 2018). A critical problem in traffic accident prevention is developing an outstanding traffic accident risk prediction system. If we can predict the chance of a traffic accident in a certain region, we may use such data to alert neighboring drivers or urge them to choose a less hazardous path. However, estimating the probability of a traffic accident is difficult owing to the many factors that might impact a collision. The incidence of traffic accidents, for example, varies substantially across localities. Furthermore, bad weather, such as snow or fog, may reduce road visibility and traffic capacity, raising the risk of traffic accidents. The rate of traffic accidents fluctuates during the day, according to the physical condition of the drivers. Even though many researchers have focused on identifying key features associated with traffic accidents, reliable adaptive prediction of traffic accident risk remains a challenge. Two sorts of methodologies are available for real-time crash risk prediction: statistical and learning methods. The conditional logit model, the log-linear model, and logistic regression are all statistical analysis methods. (Ren et al., 2018). These models are often based on matched-case control data and include a number of assumptions. (Wang et al., 2015; Yu et al., 2016). Meanwhile, Machine learning approaches such as Support Vector Machine (SVM) (Yu and Abdel-Aty, 2013), Random Forest (Lin et al., 2015), and others have grown popular as a result of these constraints. However, the majority of these forecasting systems relied on shallow traffic models, which were deemed unsuitable for large data scenarios (Nguyen et al., 2018). On the other hand, with the rapid growth of deep learning, it has recently been used to tackle a variety of transportation issues. The existence of vast data human-induced in metropolitan settings led to an increase in human movement research (Bao et al., 2017; González et al., 2008; Hasan et al., 2013). In areas like traffic flow prediction and travel demand projection, multi-sourced big data may be extremely beneficial to transport networks. Big data might hypothetically be used in road safety research to aid road experts in better understanding the mechanism and contributing causes of collisions (Bao et al., 2019). The primary benefit of Artificial Intelligence (AI) techniques is that they can predict possible real-time crashes without the need for a predetermined mathematical model. Since the artificial neural network method learns, organizes, and adapts itself through a network of nodes – like neurons inside the human brain – it is more effective for prediction. Deep learning algorithms produce excellent outcomes in various machine sensing, sight, speech, and signal processing applications. Today, Deep learning algorithms are the most reliable and applicable prediction methods. Deep neural networks or deep neural learning can learn from many unstructured and unlabeled data. Deep learning has been progressively completed in response to the massive rise of big data worldwide, dubbed "big data." It would take years, if not decades, to gather and analyze enormous amounts of data from different sources worldwide. Deep learning is used to learn this massive volume of unstructured and unlabeled data ([Alafi, 2019](#)). With the fast growth of artificial intelligence and deep learning technologies, more complex algorithms such as recurrent neural network (RNN), convolutional neural network (CNN), and long short-term memory (LSTM) or a combination of these algorithms such as LSTM-CNN are becoming accessible. Using a convolutional neural network with long short-term memory, ([Li et al., 2020](#)) created a real-time crash risk prediction model for arterials (LSTM-CNN). Weather conditions, traffic flow patterns, and signal timing were among the elements that the suggested technique learned from. Experiments revealed that the suggested model outperformed linear approaches. ([Cai et al., 2020](#)) Applied the deep convolutional generative adversarial network (DCGAN) model, an upcoming deep learning technology, to completely grasp the traffic data that leads to accidents. ([Ren et al., 2018](#)) Gathered large amounts of traffic accident data, examined the geographical and temporal patterns of traffic accident frequency, and then provided the spatiotemporal correlation of traffic accidents. They suggested a high-accuracy deep learning model-based RNN for predicting traffic accident risk based on the patterns discovered in the research. ([Wang et al., 2015](#)) studied the predicting the risk of crash on different areas including crash prediction for expressway weaving segments. ([Formosa et al., 2020](#)) conducted research on unified infrastructure and used Deep Learning to forecast traffic issues. To construct the model, highly disaggregated traffic data and in-car sensor data from an instrumented vehicle were collected from a portion of a highway. A Regional–Convolution Neural Network (R-CNN) model that recognized lane lines and tracked cars from pictures collected by a single front-facing camera was used to identify traffic issues. Using a centralized digital infrastructure, this data was combined with traffic factors and estimated safety surrogate measures to create a set of Deep Neural Network models to forecast traffic conflicts. The findings revealed that the time to collision was affected by traffic density, speed, and weather. CNNs are very powerful in sequence learning and may be more appropriate for predicting real-time crash risk by considering time series characteristics. Real-time traffic flow prediction has enabled consumers to escape traffic jams by choosing less crowded routes thanks to massive traffic data and deep learning. Big traffic data and deep learning could help predict or reduce the danger of traffic accidents. Methods based on deep learning and extensive data have demonstrated promising results in traffic-related issues, including traffic flow prediction, arrival time estimate, origin-destination prediction, and so on, as deep learning has progressed. Traffic accidents, in particular, may be linked to the day of the week or other relevant elements such as weather, air quality, and driver status. To increase the accuracy of traffic accident risk prediction, all of these parameters must be included in a complete model ([Ren et al., 2018](#)). Nonetheless, a few researchers have used deep learning approaches to analyze traffic safety. In light of these research gaps, this study aims to study the use of deep learning approaches to forecasting real-time crash risk on urban arterials. Our primary purpose in this paper is to train a system using the available traffic data to forecast the probability of a crash in future time steps to enable drivers to avoid areas where the crash risk is considered high. This paper has made several advances in predicting real-time crash risk on the urban highway and is considered one of the first to investigate using CNN in real-time crash risk prediction in Turkey. Second, the prospects of using multiple data sources, such as average vehicle speed, average vehicle volume, and weather data, for real-time crash prediction are investigated. A year's worth of data for the "TEM" highway, which is one of the main and mostly used arterials in Istanbul, is gathered from Istanbul metropolitan municipality. Thirdly, this data is pre-processed in three 4-minute intervals (12 minutes) prior to any observed crash ([Yuan and Abdel-Aty, 2018](#)). One number is implied as the crash risk for each 4-minute interval, which we have already considered as time duration prior to a crash, specifying the probability of a crash occurring in the upcoming 4-minute intervals. The same data is provided at the same time on the mentioned highway on another day without a crash. Then this gathered data is used as the training and test data to train a deep learning network. Finally, the performance of the implemented network is reported as prediction performance.

## 2. Methodologies

Basic neural network, CNN which is one of the most-known methodologies of the deep neural network, and their performance factors are explained in this section.

2.1. Deep Neural Network

Deep Neural Network (DNN), also known as Deep Learning (DL), is an AI method. Artificial intelligence is a branch of computer science that tries to replicate or copy human behavior on a machine, allowing machines to do tasks that would otherwise need the use of a human's intellect and abilities. AI systems have programmable features such as organizing, remembering, comprehending, problem-solving, and decision-making. Algorithms that employ techniques like machine learning, deep learning, and applied logic help artificial intelligence systems. Learning algorithms organize data and prepare it for usage in AI systems using statistical algorithms. Without any particular preparation, learning techniques enhance AI systems (Alafi, 2019). AI overarch machine learning and deep learning, as it is shown in Fig. 1.

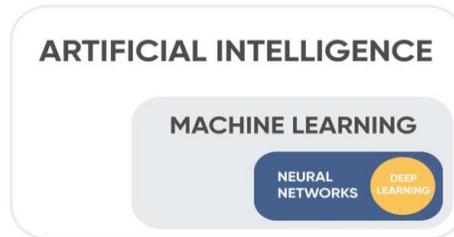

**Fig. 1.** AI and learning Methods.

Machine learning (ML) is a type of human-machine interaction that gives systems intelligence and allows them to learn from humans and act like them without requiring any particular mathematical programming. Deep learning is a type of machine learning technology. Because it can learn from a large amount of disorganized and unidentified data – known as big data – in the unsupervised category, it is referred to as a deep neural network or deep neural learning. Deep learning has progressed in reaction to the massive data boom in every area today, dubbed "big data." Gathering and analyzing massive amounts of data from many sources worldwide would take a long time, maybe decades, for a person. Deep learning is applied to learn this vast amount of unstructured and unlabeled data. Deep learning has two tasks: regression and classification. *Regression* is a prediction in which the expected aspect is a quantity and value of a single variable.
In contrast, classification is a type of prediction in which the anticipated aspect is a class or category of that variable. Multi-Layer Perceptron (MLP), Convolutional Neural Networks (CNN), and Recurrent Neural Networks (RNN) are the three most well-known deep learning techniques. MLP is the most basic and traditional DL model. MLP uses a simple neural network model with one input layer, one or more hidden layers, and one output layer for mapping from input to output. The architecture of one simple MLP as a neural network is represented in Fig. 2.

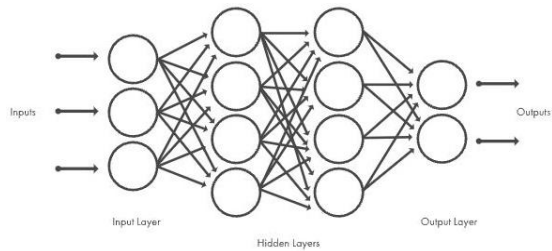

**Fig. 2.** Neural Network Structure.

All deep learning models have two training phases: feed-forward and back-propagation (BP). Deep learning employs the modification of weight values as a learning strategy. BP, sometimes known as "backward error propagation," is a strategy for updating weights and training multiple neural networks or deep neural networks, according to Goodfellow et al., 2016. In the backpropagation process, the mistakes are sent to hidden layers to train the network. Weight cannot be changed if the mistake is not well learned and does not reach the buried layers. Any neuron has an activation function, which can assist with the training problem. In general, the neural network mathematical method may be described as follows:

First step - weight initializing.

Second step - error calculation as (e), which can be by the mean squared error (MSE).

$$e = \frac{1}{N}\sum_{i=1}^{N}[t_i - o_i]^2 \quad (1)$$

Third step – reducing error by adjusting weights as follow:

$$w(n) = -\eta\frac{\partial e}{\partial w} + \alpha w(n-1) \quad (2)$$

$$\Delta w = -\eta\frac{\partial e}{\partial w} \quad (3)$$

Where, w stands for weight $\eta$ and $\alpha$ are constant. The weight is adjusted according to the difference between weighted value and network output, specified as error. All above mentioned procedures are common neural network steps. According to model which is selected the other steps will be affected. Structure of weight adjustment as BP procedure, which is one of the important parts of neural network, is represented in Fig. 3.

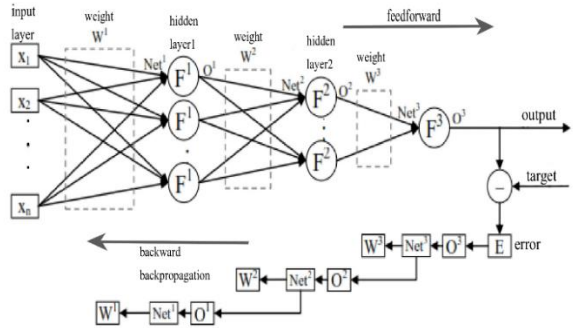

**Fig. 3.** Backpropagation procedure (Alafi, 2019).

## 2.2. CNN

Convolutional Neural Networks are another essential and widely used deep learning technique (CNN). According to the following definition, convolution is a mathematical process that moves a function along its dimensions:

$$x[n] * h[n] = \sum_{k=-\infty}^{+\infty} x[n-k].h[k] = \sum_{k=-\infty}^{+\infty} x[n].h[n-k] \quad (4)$$

$$(x * h)(t) = \int_{-\infty}^{+\infty} f(\tau)g(t-\tau)d\tau = \int_{0}^{t} f(\tau)g(t-\tau)d\tau \quad (5)$$

As there is a convolution operation, a random filter is used to scan a random input according to its dimension and provide this convolution operation. the convolution operation is given in Fig. 4.

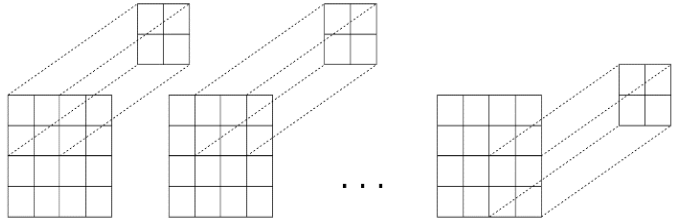

**Fig. 4.** Convolution operation (Alafi, 2019).

To get the optimal result, the filter will convolve across each filter and obtain the spatial feature of the data. The three essential layers of a CNN are the convolutional layer, the pooling layer, and the fully connected layer. The convolutional layer receives data, modifies it, and then passes it on to the next layer. During the pooling layer, a function based on the maximum operation is applied to a portion of the input to create an output. The pooling layer's output is dimensionally more petite than the input. A fully connected layer with a neural network construction stays behind the convolution and pooling layers, and features and representations learned from data via the convolutional and pooling layers are fed to this fully connected neural network without being hand-engineered, as is the case with machine learning. A CNN can have one or more convolutions and a pooling layer. In other words, the pooling layer's output represents the artificial neural network's future input. Fig. 5 illustrates the architecture of a CNN in a simple way.

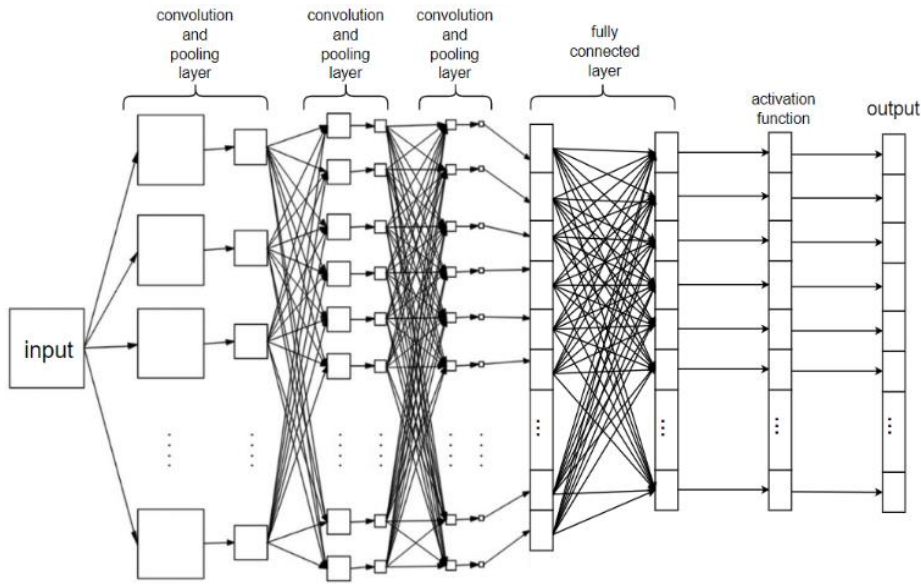

**Fig. 5.** CNN (Alafi, 2019).

One-dimensional (1D), two-dimensional (2D), and three-dimensional (3D) CNNs are classified by their input and output dimensions (3D). This category was created based on the dimensions of the provided data. For time series with one-dimensional data, a one-dimensional CNN is employed, however the CNN layer is two-dimensional since the CNN will contain input characteristics as one dimension and time step as another dimension. Two-dimensional CNN is utilized for picture data, but with a three-dimensional CNN layer. 3D CNN with four dimensional CNN is utilized or three-dimensional images such as MRI or CT scan. We will utilize 1D-CNN since this research is about time series, and time series are one-dimensional data sets. The architecture of the proposed 1D-CNN is shown in Fig. 6.

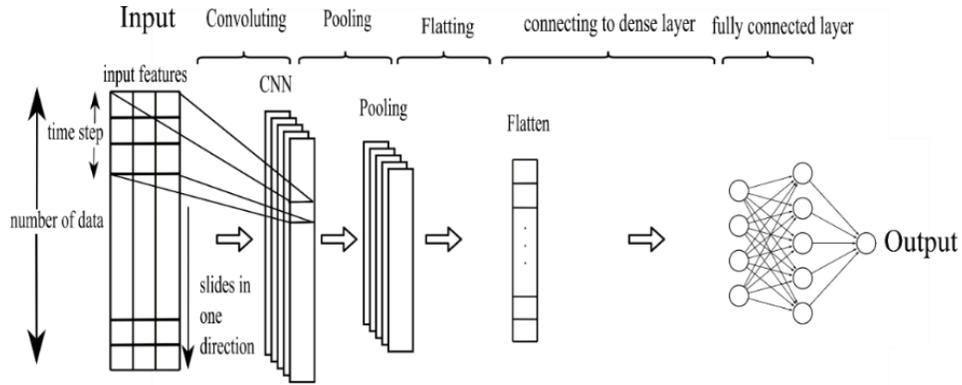

**Fig. 6.** 1-D CNN architecture.

The convolutional layer's filters first convolve the input; then, it travels through the pooling layer to be smaller in dimension than the input. Convolution and pooling layers are two-dimensional, and a CNN may contain many of them. The output of the last pooling layer is then flattened. The output of the pooling layer is a 2D feature map, but the input of the fully connected layer is a 1D feature vector that may be used as a scalar value. The pooling layer's 2D output is converted to a fully connected layer's 1D input via flattening. The learning process is then completed, and the processed data is input into a neural network, which generates the results. The dimensions of the input data are crucial in 1D-CNN. The input is a 2D matrix in which the characteristics of the input data are one dimension, and the length of the data set is the other. These many properties have distinct values at any given moment. The time step specifies the number of time steps that will be examined simultaneously before being transferred to the CNN layer and slid along the time series. After transferring each group of data to the CNN layer by the dimension of features multiplied by time steps, CNN extracts the primary feature of each group using activation functions such as ReLu and optimization methods such as Adaptive Moment Estimation (ADAM).

*2.3. Performance Estimation*

The performance of a neural network is influenced by the method's inaccuracy, as well as a few other elements that will be covered. As previously stated, throughout the neural network process, there is an error function that describes the difference between the predicted output and the calculated result. In a neural network, the error term is very important since it is used to alter the weights. The mean squared error (MSE) is a sort of error function with the following equation:

$$MSE = \frac{1}{N}\sum_{i=1}^{N}(t_i - y_i)^2 \qquad (6)$$

Root mean squared error (RMSE) is another error function which is considered as another performance factor can be stated as follows:

$$RMSE = \sqrt{\frac{1}{N}\sum_{i=1}^{N}(t_i - y_i)^2} = \sqrt{MSE} \qquad (7)$$

The correlation coefficient, also known as the determination coefficient, is another metric symbolized by the letter R. This option specifies the relationship between the aim and the predicted value. In other words, this coefficient indicates the linearity between the actual and expected values. The equation is as follows:

$$R = \sqrt{1 - \frac{\sum_{i=1}^{N}(t_i - y_i)^2}{\sum_{i=1}^{N}(t_i - \bar{t})^2}} \qquad (8)$$

In all of the equations above, t represents the goal value or intended output, y represents the calculated or predicted output, and N represents the number of periods of time. RMSE appears to be the square root of MSE. However, there are some practical distinctions between these two numbers. The minimizers are obviously related; if a set of predictions reduces the MSE, it will also minimize the RMSE. However, the difference is in the scale; the error acquired by RMSE is on the same scale as the goal value. Its value provides an acceptable feeling of inaccuracy in the prediction process. Another accuracy index for binary classification is the receiver operating characteristic curve (ROC), a graphical representation of a binary classifier system's diagnostic performance when its discrimination threshold is modified. The ROC curve is a prominent way of conceptualizing a classifier model's behavior (Provost et al., 1997). ROC analysis provides for a more independent and comprehensive evaluation of classifier performance than just utilizing accuracy. Because it is simple to define, understand, and computationally viable, ROC analysis is commonly provided for two groups. The true-positive rate (TPR) on the y-axis and the false-positive rate (FPR) on the x-axis are shown on the ROC analysis for two classes, yielding a point for each classifier. We may get an endless number of derived classifiers along the segment that links two classifiers simply by voting them with varying weights, resulting in a "curve" (Fawcett, 2006). As a result of the lower TPR and/or higher FPR, each point below that segment will have a higher cost for any class distribution and cost matrix. Given a set of classifiers, the ones that lie inside the convex hull formed by the points representing the classifiers and the points (0,0) and (1,1), which represent the default classifiers that always predict negative and positive, respectively, can be discarded. The traditional technique of representing the ROC space, in which the true class is progressively represented for right predictions, and the false class is incrementally represented for erroneous predictions, does not seem to us to be particularly consistent. In addition, this option is difficult to extend beyond two classes. We recommend that the True positive rate (TPR) and the false-positive rate (FPR) be represented instead. The point (1,0) is called a perfect classifier which never misclassifies any positive classes. By starting from point (0,0) and changing the threshold between classes, we can proceed until the point (1,1). In this study, the area under the ROC curve (AUC) (Hanley and McNeil, 1982) is used for model evaluation. AUC is a two-dimensional measurement of the complete area beneath the whole ROC curve from (0, 0) to (1, 1). Other performance parameters such as true positive rate and false alarm rate are also presented. True Positive Rate (TPR) is a term used to describe sensitivity, whereas False Positive Rate (FPR) is a term used to describe false alarm rate (FPR). The following are the TPR and FPR formulations:

$$TPR = \frac{TP}{TP + FN} \qquad (9)$$

$$FPR = \frac{FP}{FP + TN} \qquad (10)$$

Another parameter is the positive predictive value as Specificity or Precision (Pr) which is defined as below:

$$Pr = 1 - FPR = \frac{TP}{TP + FP} \qquad (11)$$

The classification confusion matrix is used to determine the values of TP (True Positive), FP (False Positive), FN (False Negative), TN (True Negative). Table 1 represents confusion matrix for a binary classification.

**Table 1.**
Confusion Matrix

|  | True crash | True non-crash |
|---|---|---|
| Predicted crash | TP | FP |
| Predicted non-crash | FN | TN |

## 3. Data acquisition and Preparation

### 3.1. Data acquisition

This research focuses on real-time crash risk prediction on one of Istanbul's major motorways. The TEM Highway is one of Istanbul's major motorways, connecting the Asian and European sides of the city. Istanbul is Turkey's biggest metropolis, with a population of almost 16 million people; hence, traffic and traffic-related accidents are one of the city's significant worries. As previously said, the speed limit on this highway is not restricted to low numbers since it is one of the quickest routes for commuters to move from one side of the city to the other. As a result, the indicated motorway has a high number of recorded collisions each year. Fig. 7 illustrates the route of TEM highway.

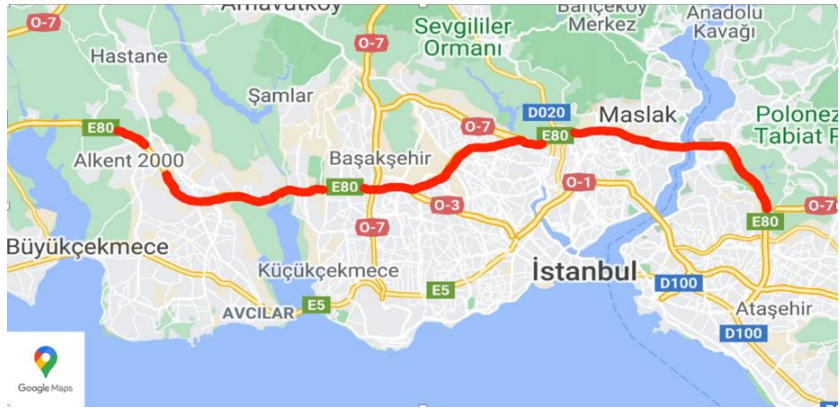

**Fig. 7.** TEM highway route.

Regarding data availability, the portion of this highway we employed in our study is about 50 kilometers long. It is equipped with 36 cameras equipped with sensors that offer the necessary data for our research. The placement of the cameras is shown in Fig. 8, which was generated using Google Maps. From the 1st of January 2019 to the 31st of December 2019, five types of data were collected, including the average speed of vehicles on each side of the freeway, the volume of vehicles on each side, the volume of cars in the entry lanes, the temperature of each day, and precipitation data, all of which were provided by The Istanbul Metropolitan Municipality.

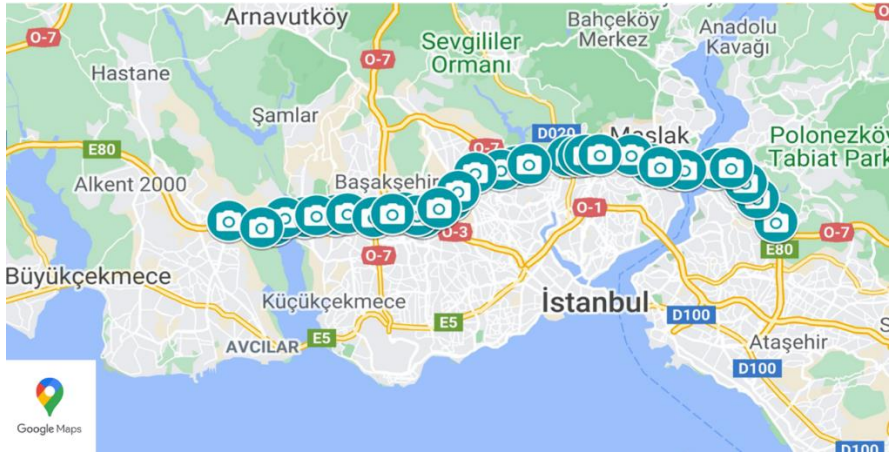

**Fig. 8.** Sensors' locations.

This data is recorded every second by the 36 cameras along the road. All data is considered as raw data. This data requires to be preprocessed before entering the deep learning system. Approximately eight million data are recorded in a single year with around one thousand and three hundred crash events. The rarity of the accident occurrence case characteristic is one of the primary obstacles in effectively predicting real-time traffic crash risk. In actual traffic conditions, non-crash instances vastly outweigh crash cases in experimental observations, resulting in an unbalanced data categorization challenge in accident risk analysis (Yehia et al., 2021). Since the number of accident instances in this research relative to non-crash cases is very low, crash cases are regarded as outliers and worthless data among all data. As a result, as previously said, data preparation is critical. To preprocess, we first take one accident that has been documented and record all five kinds of data at the time of the occurrence as our crash data, and then we do the same for another day at the identical time when no crash has been reported as our non-crash data. Finally, the accident and non-crash data are separated into three 4-minute periods, allowing us to forecast the chance of a crash in the next four minutes.

3.2. Data Preparation

As previously stated, there are around eight million recorded traffic data for the TEM highway in 2019 that contain both accident and non-crash data, with 1293 data for crashes and the remainder for non-crash occurrences. The ratio of a crash to a non-crash occurrence is around 1:6152, indicating that our data is skewed. The crash prediction model performs poorly when we use the original data with no alterations. Data resampling is a methodical approach to tackling this problem. To deal with this and minimize the ratio, we selected three to four minutes of data prior to the precise time of the recorded accident on the same day and at the identical geographical locations. Furthermore, we repeated the process on another day when no accident was reported, using the same time and locations but on a different day. (Liu and Chen, 2017). The first set of data was utilized as our crash, and the second as our non-crash data. In this manner, we can analyze the data and therefore lower the crash-to-non-crash data ratio to 1:5, including 7758 data, of which 1293 is for the crash and the rest is for non-crash data. Table 2 represents Statistical characteristics of recorded data. We picked four minutes since the cameras installed on this roadway capture data every four minutes. Fig. 9 shows the time slices in which our data has been selected before actual crash time.

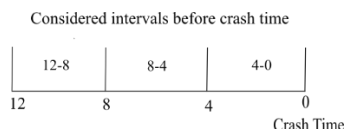

**Fig. 9.** Crash Labeling

**Table 2.**
Statistical characteristics of recorded data.

| Type | Feature | Description | Mean (std) | (Min Max) |
|---|---|---|---|---|
| Traffic Data | up-speed | Average speed in upstream | 75.06 (26.23) | (3 123) |
| | down-speed | Average speed in downstream | 72.49 (29.50) | (4 247) |
| | V1,2,3,4 | Average volume of upstream | 176.52 (57.40) | (1 314) |
| | V5,6,7,8 | Average volume of downstream | 113.78 (61.95) | (0 296) |
| | VL1 | Volume of vehicles entering from lane 1 | 12.63 (16.50) | (0 246) |
| | VL2 | Volume of vehicles entering from lane 2 | 16.19 (19.84) | (0 207) |
| | VL3 | Volume of vehicles entering from lane 3 | 16.41 (23.34) | (0 93) |
| | VL4 | Volume of vehicles entering from lane 4 | 14.42 (20.88) | (0 88) |
| | VL5 | Volume of vehicles entering from lane 5 | 13.70 (18.02) | (0 105) |
| | VL6 | Volume of vehicles entering from lane 6 | 11.16 (12.08) | (0 68) |
| | VL7 | Volume of vehicles entering from lane 7 | 2.35 (5.43) | (0 61) |
| | VL8 | Volume of vehicles entering from lane 8 | 2.40 (5.11) | (55 0) |
| | CR | Crash risk | 25 (0.39) | (0 100) |
| Weather Data | Temperature | Average temperature of the day in degrees Celsius | 17.71 (7.00) | (0 29) |
| | Precipitation | No rain: 0<br>Rain: 1 | 0.17 (0.38) | (0 1) |

### 3.2.1. Feature Selection

Pearson correlation coefficients and Extra-tree (Geurts et al., 2006) were used to investigate feature correlation and importance (Benesty et al., 2009). Fig. 10 illustrates the conclusions of feature importance and feature correlation. Following the stage, all features are normalized regarding their maximum and minimum values. A meta-estimator is used to fit a variety of randomized decision trees on distinct sub-samples of the dataset, and then averaging is used to prevent over-fitting and increase anticipated accuracy (Li et al., 2020, Pedregosa et al., 2011). The Pearson correlation coefficient, which ranges from -1 to 1, measures the relationship between each pair of characteristics. It reflects the degree to which two variables are related. If the absolute value of a correlation is more than 0.5, it is termed strong (Cohen, 1992; Li et al., 2020). If one feature has a substantial correlation with another trait but is less valuable than the other, it will be removed. After implementing the feature selection method, according to the importance and correlation factor for variables of this study, the complete dataset had ten features with the inclusion of up speed, down speed, upstream volume, downstream volume, volume entries 2,3,5 and 6, temperature, and precipitation.

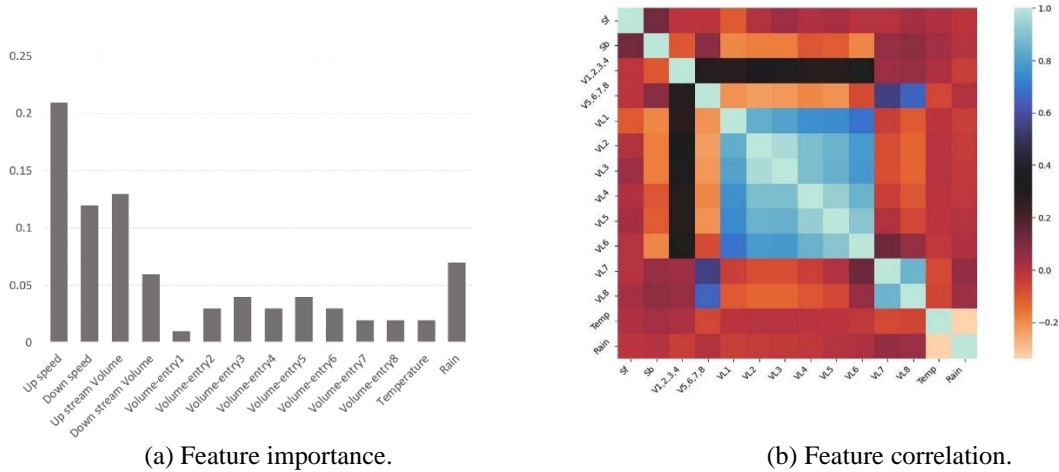

(a) Feature importance.    (b) Feature correlation.

**Fig. 10.** Feature importance and correlation

## 4. Experiments Design and Performance

*4.1. Exploited Network Architecture*

As previously indicated, we will choose our deep learning type based on the kind of data we have and the eventual result we want from this network. In this study, we are to identify future crash risk based on 10 meteorological data points that we have; therefore, it is regarded a time series prediction issue; we may use LSTM or one-dimensional CNN, or we can use a combination approach like CNN. In this case, we employed one-dimensional CNN to forecast the likelihood of a crash. The network architecture of our CNN is shown in Fig. 11. The procedure diagram of the 1D-CNN model which is used in this research is shown in Fig. 12.

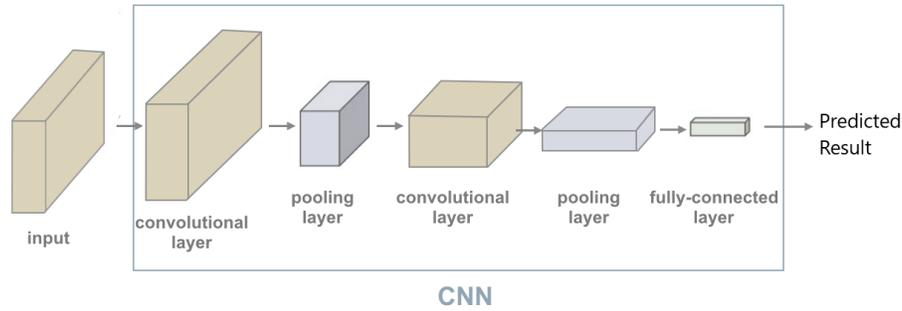

Fig. 11. Architecture of CNN.

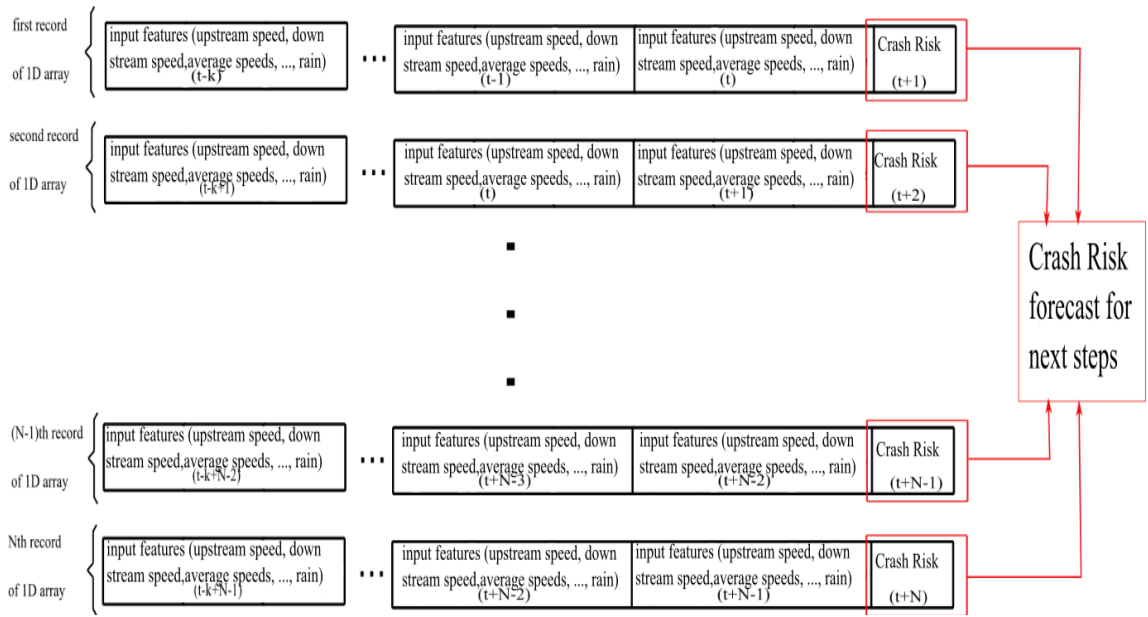

Fig. 12. 1D-CNN procedure

According to the number of input data in hand, we have ten input data as upstream speed, downstream speed, the volume of vehicles on each side of the freeway, volume of the vehicles from entry lanes, and weather; we should select multivariate CNN type, and consider these different inputs as features of the input. After that, we will enter these ten inputs under the name of a feature of input as columns of a two-dimensional input matrix and then select k rows of this matrix and apply them to the CNN layer. K represents time steps or sampling parameters. Training will be done through this 1D-CNN network, and finally, we will learn to find crash risks a step ahead. Now, we have one multivariate 1D-CNN. Input data is two-dimensional in this CNN, and its dimensions stand for features mentioned above, plus time steps as sampling time by the meaning of steps we want to sample from these features. In general, it is like a matrix in which some rows stand for time steps, and the column stands for input features. Rather than having time steps, the data is treated as a sequence over which the CNN may perform convolutional read operations, similar to a two-dimensional image. In order to train the model, we will have to ignore specific values from the output time series when we do not have values in the input time series at previous time steps. As a result, the number of input time steps chosen significantly impacts how much of the training data is utilized. After this sampling method from input data and creating a two-dimensional matrix with these ten one-dimensional separate time series, sampled data is routed to the convolutional and pooling layer and then flattened layer to reduce the dimension of the feature map to a one-dimensional series or vector to be able to be transferred to dense or fully-connected layer and after that be obtained as output. We considered 80% of the data for training and the other 20% of it for the test of the algorithm. To get the best result, one of the most important steps in the training process is to choose the right hyperparameters and optimization algorithms. The base of implemented model is NVIDIA GEFORCE 940 MX 8G CPU. Moreover, the model's module's filter and batch size, epoch number, and learning rate are modified, as shown in Table 3.

**Table 3.**
Parameters tuning.

| Hyperparameter | Value |
| --- | --- |
| Learning Rate | 0.01 |
| Epoch Number | 100 |
| Batch Size | 10,000 |
| CNN Filter Size | 64 |

*4.2. Experiment Performance*

After sampling the data and dividing data into 80% as training data and 20% as testing, the proposed CNN model is trained based on training data; then, the trained model is tested on testing data. The results of the proposed model are presented as follows. Fig. 13 shows observed crash risk values vs. predicted crash risk values for all data (training and testing) as one of the model results in which the blue section represents observed (actual) crash risk while the red section represents predicted crash risk. The correlation factor between predicted and observed values is obtained as 0.9467.

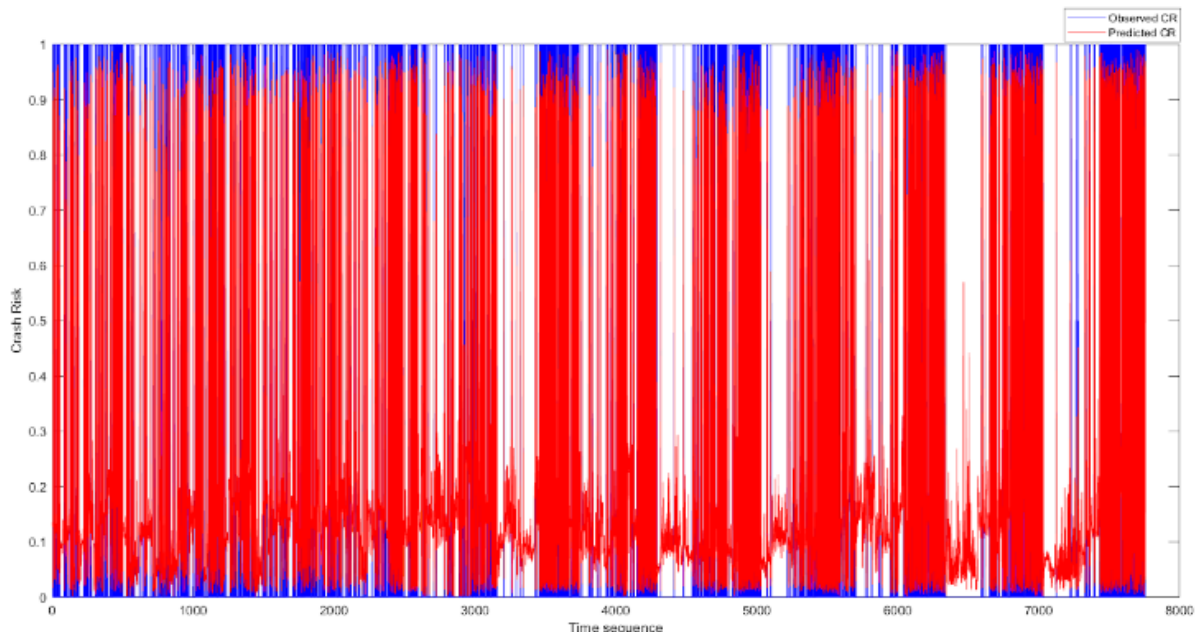

**Fig. 13.** Predicted and observed crash risk value

Fig. 14 shows changes in error value (MSE) during the execution of the CNN model as another performance factor of the model.

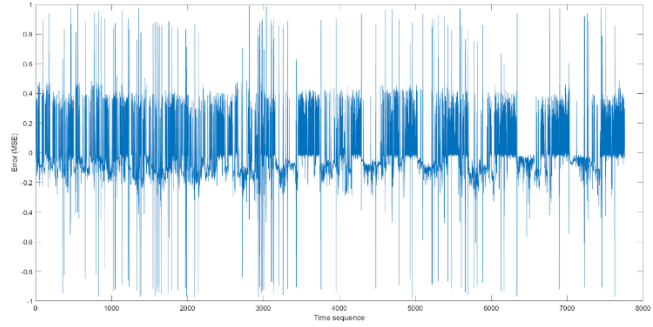

**Fig. 14.** MSE value changes.

The ROC curve is commonly utilized in binary classification issues based on drawing the FPR and TPR at various thresholds (Gunčar et al., 2018). In this study, as we have three classes, no crash risk, low risk, and high risk, which were defined as 0, 0.5, and 1 according to their crash likelihood, we cannot use the ROC curve directly. In multiclass classifications, we should split it into multiple binary classes and use averaging ROC to check the performance of our model's prediction (Sokolava et al., 2009). Macro and micro averaging are two methods that are used for averaging ROC. Macro and micro averaging methods and their results have slight differences from each other. In this study, the results of micro and macro averaging are represented. in the macro averaging method, first of all, precision for three binary classifications as one vs rest is calculated, then the average of three different precisions is calculated as macro precision as follows:

$$Pr_{macro} = \frac{Pr_1 + Pr_2 + Pr_3}{3} \qquad (12)$$

where $Pr_i$ is the precision obtained by recoding the multiclass predictions to only class i. When contributing their fraction of the accuracy value to the whole (in this case, 1/3), all categories gain equal weight in macro averaging. As there is a significant level of imbalance data, this may not be a reasonable assessment. In such a situation, a weighted macro average, with weights determined by the frequency of that class in the truth column, could be preferable.

$$Pr_{weighted-macro} = Pr_1 \frac{\#Obs_1}{N} + Pr_2 \frac{\#Obs_2}{N} + Pr_3 \frac{\#Obs_3}{N} \qquad (13)$$

Where $\#Obs_i$ stands for the number of observations for class i, and N stands for the total number of observations. As micro averaging method, it considers the full collection of data as a cumulative result, and instead of calculating three discrete precision metrics, it produces just one precision metric as average. To achieve accuracy, it calculates all true positive results for each class and uses it as the numerator, then calculates all true positive and false positive results for each class and uses that as the denominator which are all derived from the confusion matrix.

$$Pr_{micro} = \frac{TP_1 + TP_2 + TP_3}{(TP_1 + TP_2 + TP_3) + (FP_1 + FP_2 + FP_3)} \qquad (14)$$

Instead of each class being given equal weight, each observation is given equal weight in this situation. This increases the potency of the classes with the greatest observations. Therefore, parameters of micro averaging methos is considered as accuracy factors in this study. The ROC curve's threshold is selected at the point where true positive rate known as sensitivity equals true negative rate known as specificity or precision to measure the precision or sensitivity and false alarm rate. ROC curve and performance parameters for implemented model are represented in figure 15 and table 4.

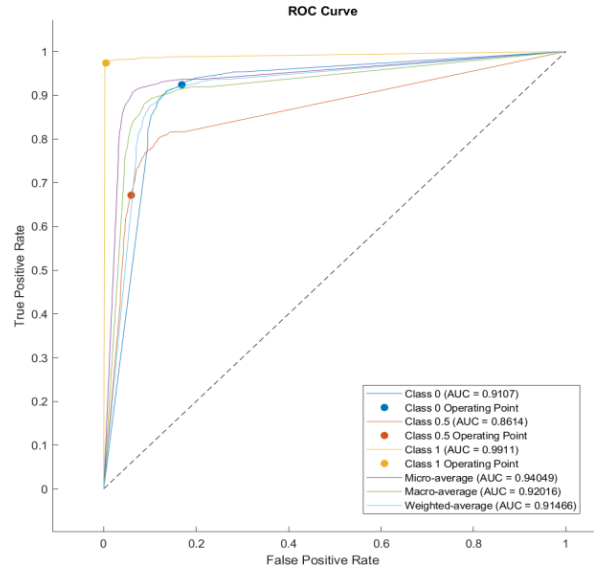

**Fig. 15.** MSE value changes

**Table 4.**
model's results for three classes.

| Metric name | AUC | False alarm Rate | Precision | MSE | RMSE | R |
|---|---|---|---|---|---|---|
| Class 0 | 0.9107 | 0.12 | 0.88 | | | |
| Class 0.5 | 0.8614 | 0.18 | 0.82 | | | |
| Class 1 | 0.9911 | 0.05 | 0.95 | 0.024 | 0.1549 | 0.9467 |
| Micro | 0.94049 | 0.08 | 0.92 | | | |
| Macro | 0.92016 | 0.11 | 0.89 | | | |
| Weighted-macro | 0.91466 | 0.12 | 0.88 | | | |

With a high correlation factor, high AUC value, high sensitivity, a low error value, and a low false alarm rate the suggested model provides delicate outcomes. The performance parameters of implemented model are shown in table 5.

**Table 5.**
The model's performance.

| Metric name | AUC | False alarm Rate | Precision | MSE | RMSE | R |
|---|---|---|---|---|---|---|
| Value | 0.94049 | 0.08 | 0.92 | 0.024 | 0.1549 | 0.9467 |

These three occurrences as crash likelihood are obviously hard to separate in the raw characteristics, and their patterns are intertwined. The traits retrieved from CNN, on the other hand, successfully separate them. The crash likelihood as three occurrences become nearly indistinguishable. It's worth noting, though, that all occurrences are still mixed in with each other. The performance of the model can be elevated by considering some factors in the future.

## 5. Model Comparison

The performance of the aforesaid CNN model is compared with three selected neural network and machine learning benchmark models with the inclusion of the Backpropagation neural network, Support Vector Machine (SVM), and Decision tree. One BP neural network and two machine learning models are trained according to the same input dataset for predicting the same output data, and the results of them are compared to the supposed CNN model. Table 6 and Figure16 represent the performance parameters for the proposed CNN model and three compared models. Results of the comparisons. The suggested CNN has the best AUC value of 0.94049, outperforming the other models. Furthermore, Fig. 16 shows that the suggested CNN has the maximum sensitivity or precision and R (Correlation Coefficient) and the lowest false alarm rate and error value, demonstrating its superior performance from another point of view. To conclude this comparison, the suggested CNN surpasses existing approaches in terms of AUC, sensitivity, false alarm rate, error, and correlation factor. The findings support the practicality and superiority of the proposed technique, which can adequately estimate accident risk on arterials in real time.

Furthermore, one of the suggested model's most distinguishing features is its low false alarm rate. In general, deep learning and machine learning models tend to outperform statistical models in terms of accuracy. The primary purpose of this paper is to show the feasibility and application of executing deep learning methods in multiclass crash risk prediction.

**Table 6.**
Different model's performance parameters.

| Metric name | AUC | False alarm Rate | Precision | MSE | RMSE | R |
|---|---|---|---|---|---|---|
| CNN | 0.94049 | 0.08 | 0.92 | 0.024 | 0.1549 | 0.9467 |
| BP | 0.8512 | 0.18 | 0.82 | 0.05 | 0.2236 | 0.8273 |
| SVM | 0.57 | 0.35 | 0.65 | 0.145 | 0.38 | 0.65 |
| Decision tree | 0.82 | 0.21 | 0.79 | 0.046 | 0.215 | 0.82 |

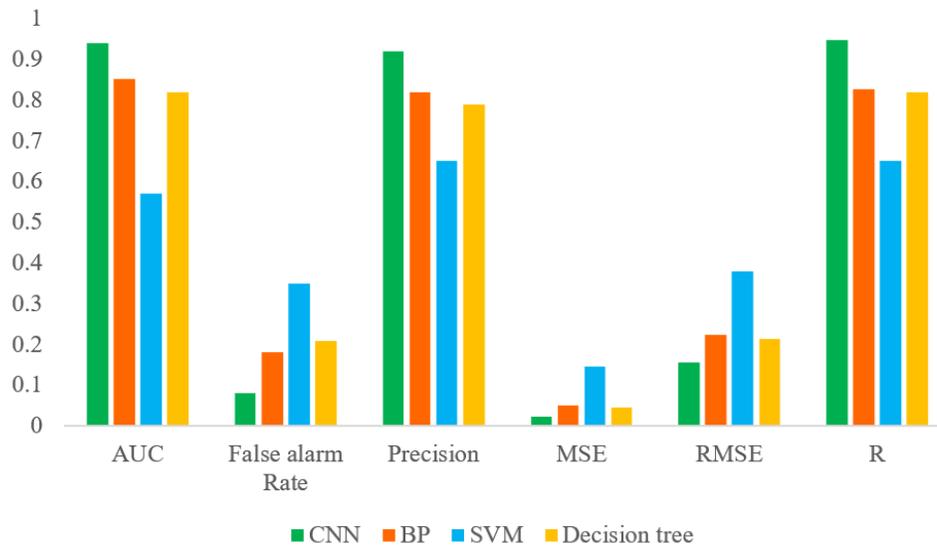

**Fig. 16.** Model's performance comparison

## 6. Conclusion

This research used a convolutional neural network to forecast real-time crash risk on one of Istanbul's major motorways. First of all, various data such as average speed, vehicle volume in each lane of highways, the volume of cars in entrance lanes, precipitation, and meteorological conditions were gathered for a year. Second, raw data were pre-processed and balanced based on Pearson correlation and importance factors. Thirdly we recorded the traffic data for three 4-minutes before the actual crash as our crash data, and we applied the same procedure on other days in which no crash had been recorded as non-crash data. Finally, a deep neural network was developed to forecast real-time crash risk, and three neural network and machine learning methods were used compared to the implemented CNN model. One multivariate 1D-CNN was trained to trace and learn features and crash risk behavior. Performance factors of the implemented network were reported and were in the acceptable range. Changes and differences between AUC and the other performance parameters of the ROC curve were also discussed to interpret the numerical results and explain the network's performance. The findings show that the suggested CNN outperforms the others in various ways.

To begin with, it has a maximum sensitivity of 92 percent, and correlation factor of 94.67 percent, and the lowest false alarm rate and error of only 8 percent and 1.549 percent, respectively. Moreover, it has the most significant AUC of 0.94, which is much greater than other approaches such as BP, SVM, and Decision tree models. The findings of this study support the practicality and superiority of the proposed technique, which can properly estimate accident risk on arterials in real time. As a result, this study represents that CNN described above can be trained and predict one step ahead crash probability according to classification, which we have defined as No, low, and high risk with the probability of 0, 0.5, and 1, respectively, according to this data. The findings of this study can be utilized to develop an improved traffic management system that has the ability to lessen collisions by warning drivers. Despite the advantages of this study, there are some drawbacks. First, the performance of the implemented model may be enhanced by adding additional layers or experimenting with new hyperparameter combinations.

Furthermore, geographical characteristics might be regarded as feasible input data for the implemented model, which may aid in the performance improvement of the model. In future studies, other data, such as the characteristics of the road or even the vehicle, could be applied as our input data to maintain better outcomes. Secondly, we can consider a different range of time steps prior to a crash in order to study the effect of more time step data by considering more classes for crash likelihood. As a practical result, these results can be introduced to the transportation sector as one approach to predicting crash risk and getting ready for future plans of smart transportation systems.